\documentclass[twoside,11pt,preprint]{article}

\usepackage{blindtext}

\usepackage{caption}
\usepackage{subcaption} 

\usepackage[abbrvbib, preprint]{dmlr2e}

\usepackage[noabbrev]{cleveref}

\usepackage{lastpage}
\usepackage{graphicx}
\graphicspath{ {./images/} }

\usepackage{algorithm}
\usepackage{algpseudocode}
\usepackage{acronym}    
\usepackage{listings}
\usepackage{xcolor}
\usepackage{footnote}

\definecolor{codegreen}{rgb}{0.1,0.7,0.1}
\definecolor{codegray}{rgb}{0.9,0.9,0.9}
\definecolor{codepurple}{rgb}{0.68,0.1,0.92}
\definecolor{backcolour}{rgb}{0.95,0.95,0.92}

\lstdefinestyle{mystyle}{
    backgroundcolor=\color{backcolour},   
    commentstyle=\color{codegreen},
    keywordstyle=\color{magenta},
    numberstyle=\tiny\color{codegray},
    stringstyle=\color{codepurple},
    basicstyle=\ttfamily\footnotesize,
    breakatwhitespace=false,         
    breaklines=true,                 
    captionpos=b,                    
    keepspaces=true,                 
    numbers=left,                    
    numbersep=5pt,                  
    showspaces=false,                
    showstringspaces=false,
    showtabs=false,                  
    tabsize=2
}

\ShortHeadings{Traffic Destinations Prediction with GAMs}{Yousif and Müller}
\firstpageno{1}

\begin{document}

\newacro{SDD}[SDD]{Stanford drone dataset}
\newacro{InD}[InD]{Intersection Drone Dataset}
\newacro{EBM}[EBM]{Explainable Boosting Machines}
\newacro{GAM}[GAM]{Generalized Additive Models}
\newacro{NAM}[NAM]{Neural Additive Models}


\title{Efficient and Interpretable Traffic Destination Prediction using Explainable Boosting Machines}
\author{\name Yasin Yousif \email yy33@tu-clausthal.de\\
       \addr Department of Informatics\\
       Clausthal University of Technology\\
       Clausthal-Zellerfeld, Germany
       \AND
       \name Jörg P. Müller \email jpm@tu-clausthal.de\\
       \addr Department of Informatics\\
       Clausthal University of Technology\\
       Clausthal-Zellerfeld, Germany}


\maketitle

\vspace*{-\baselineskip}
\begin{abstract}


Developing accurate models for traffic trajectory predictions is crucial for achieving fully autonomous driving. Various deep neural network models have been employed to address this challenge, but their black-box nature hinders transparency and debugging capabilities in a deployed system. Glass-box models offer a solution by providing full interpretability through methods like \ac{GAM}. In this study, we evaluate an efficient additive model called \ac{EBM} for traffic prediction on three popular mixed traffic datasets: \ac{SDD}, \ac{InD}, and Argoverse. Our results show that the \ac{EBM} models perform competitively in predicting pedestrian destinations within \ac{SDD} and \ac{InD} while providing modest predictions for vehicle-dominant Argoverse dataset. Additionally, our transparent trained models allow us to analyse feature importance and interactions, as well as provide qualitative examples of predictions explanation. The full training code will be made public upon publication.

\end{abstract}


\section{Introduction}


Traffic trajectory prediction is the problem of predicting the future path of a traffic entity for a specific time horizon based on its current state, and it has been widely studied in public traffic benchmarks. Many previous works \cite{acvrnn,amenet,goalsar} have been evaluated on these benchmarks under standard conditions to assess their prediction accuracy. By accurately predicting traffic trajectories, we can anticipate and avoid accidents or unsafe situations on the road, which is particularly useful for self-driving cars navigating safely.

Early methods to address this problem relied on rule-based approaches that applied general rules to determine future movements, such as social force models \cite{sfm} or cellular automata models \cite{cellular}. However, these methods lack the ability to extract data-based patterns from traffic data. Later, deep learning-based models emerged \cite{sociallstm,socialgan}, which automatically extracted rules from the datasets and outperformed other methodologies on well-known traffic benchmarks such as Argo \cite{argo} and InD \cite{ind}.

In recent years, some deep learning models \cite{sophie,nsp-sfm,chauffeurnet} have focused on enhancing the explainability of their models in a post-hoc manner. This is because a black-box model that provide accurate predictions using hidden processes for feature extraction and prediction inference can decrease trust in the model and make debugging any errors difficult.

This motivated works like \cite{sophie,chauffeurnet} to extract attention maps of the input images or works like \cite{nsp-sfm} to use a more interpretable input features. However, these remain partial solutions, as they still considered an approximation of the true features importance \cite{gam1}.    

An alternative approach is to use glass-box models such as linear regression, decision trees, or \ac{GAM} \cite{gam} that provide transparent explanations at the cost of lower predictive ability. Among these methods, \ac{GAM} has been found to offer the best predictive results in some applications (e.g., \cite{healthgam,gam1}). Recent variants of \ac{GAM}, such as \ac{NAM}  or boosted trees (\ac{EBM} \cite{interpretml}), have also shown competitive prediction results while maintaining transparency. In this work, we use an \ac{EBM} model to predict traffic trajectories and evaluate our results on three well-known datasets in the field. We also extract feature importance values and partial dependence graphs that provide a clear description of how the model arrives at its predictions. These graphs show each input feature's contribution to the output variable, offering an exact representation of the model's decision-making process

In \ac{GAM}, multiple sub-models are used, with each accessing one or two input features. These sub-models were previously shown to perform competitively when implemented using boosted trees, known as \ac{EBM}. The implementation of \ac{EBM} in this work is based on the Interpretml Python library \cite{interpretml}. However, there are limitations to \ac{GAM}'s current formulation such as the inability to process images as input since they need to be processed collectively for useful information extraction. Additionally, having only a single output for the \ac{GAM} model is another restriction that can be addressed by using full models for each output variable to predict. This approach involves having two models for each $(x,y)$ coordinate of the predicted trajectory in this work. 

Due to the difficulty in predicting a single deterministic possible output trajectory (known as mode), prediction is done for multiple modes with one \ac{EBM} for each. The clustering and separation of these modes is an independent problem, addressed here by clustering the target variables

This work has two main contributions: 1) The evaluation of \ac{EBM} as a glass-box model for the first time by using it as a traffic destination prediction model on three popular traffic datasets, namely \ac{SDD}, \ac{InD}, and Argoverse. This provides insight into how transparent and interpretable models like \ac{EBM} can perform in real-world datasets.
2) We calculate partial dependence graphs and feature importance for all input features with respect to the output variable. These results offer insights into the interaction of features and their significance for traffic prediction, as well as show some selected examples of local explanations. The following sections outline GAM formula, the methodology containing the preprocessing steps, experimental evaluation, and finally, the conclusions.

\section{GAM Formula}

To define GAM and GA2M mathimtical formulas, let $\mathcal{D} = {(\mathbf{x}_i ,y_i)}^N_1$ be the training dataset, and $\mathbf{x}_i = (x_{i1},x_{i2}, .. ,x_{ip})$ the input vector for the target variable of $y_i$, and $x_j$ is the feature $j$ value. Therefore GA2M, \cite{healthgam}, will try to fit the dataset as the following expression:

\begin{equation}
    g(E[y]) = \beta_0 + \sum_j f_j(x_j) + \sum_{k\neq j} f_{kj}(x_k,x_j)
    \label{eq:1}
\end{equation}

Where $f_j$ is the model mapping function for feature $j$, $f_{kj}$ is for the interaction between features $(j,k)$, $g$ is the link function and $\beta_0$ is the bias parameter. This model will be completely explainable, because one can draw the exact relationship between the input feature $x_j$ and the target $y$ as 2D plot or as heatmap for the interaction between two features as in the last term.



With \ac{EBM}, in addition to fitting each feature alone, a fast search  for the most useful feature interactions in the dataset is done as in \cite{ga2m}, then the top $k$ features interactions are used in the model, where $k$ is selected using cross validation.

List of the hyperparameters of \ac{EBM} in this work are provided in \cref{app:b}.

\section{Methodology}

In our approach, we generate multiple outputs representing different modes and assign probabilities for each mode. We detail the representation of multi-modality and probability calculation in the first two subsections. Following that is three subsections to focus on input and output shape specifications for \ac{SDD}, \ac{InD}, and Argo datasets, respectively.


\subsection{Multiple Modes Output}

Most traffic prediction benchmarks require predicting multiple possible future trajectories with a probability distribution over them, which is known as multi-modal output  \cite{covernet}. This condition ease the prediction task; however, in our case, \ac{EBM} has a unimodal output, so we can train separate models for each mode to obtain multi-modal output.


However, defining these modes is another challenge.  Many approaches were used in the literature, like using a set of rule-based modes \cite{covernet}, or with an implied clustering in multiple branches output \cite{bcmode}.In this work, we manually cluster the target points using either K-means clustering algorithm for \ac{SDD} and \ac{InD} datasets (see \cref{fig:sdd_traget}) or uniform splitting over x-axis followed by y-axis to equally distribute samples across modes for Argoverse dataset (see \cref{fig:road_and_mode}).

\subsection{Outputs Probabilities}

Each of the previous outputs should be assigned a probability normalized over all the outputs. This will add flexibility to the prediction because it can give an arbitrary number of outputs and then only the most probable $K$ output will be taken. In this work, probability estimation of modes is done by finding the sum of the loglikelihood of each of the mode-level models $\mathcal{L}_1$ with the corresponding loglikelihood of its predictions under the distribution of the main uni-modal model $\mathcal{L}_2$, as in \cref{eq:2}. 

\begin{equation}
 \mathcal{L}_{mode_x} = \lambda_1 \mathcal{L}_1(\pi_{mode_x}(x), \sigma_{mode_x}) + \lambda_2 \mathcal{L}_2(\pi_{mode_x}(x), \sigma_{all_x}) 
\label{eq:2}
\end{equation}

After finding that for both of $x$ and $y$ axises, we take a weighted sum of them for each mode in \cref{eq:3}. Finally, the top 20 probable output will be taken as final. 

\begin{equation}
    \mathcal{L}_{mode} = \lambda_3 \mathcal{L}_{mode_x} + \lambda_4 \mathcal{L}_{mode_y}
\label{eq:3}
\end{equation}

The weights $\lambda_3$ and $\lambda_4$ are determined based on relative standard deviation between axes, while $\lambda_1$ and $\lambda_2$ can be experimentally set

\subsection{SDD}

In the \ac{SDD} experiments, our models predict the final point of each trajectory, which the most challenging part to predict. This allows for comparison with other methods in the \ac{SDD} benchmark using the Final Displacement Error (FDE) metric. In this work, we use pixels as the standard unit of measurement, similar to previous works such as \cite{covernet,goalsar,p2tirl}.

After loading data files from 60 scenes, they will be split according to a common splitting standards as in \cite{p2tirl}, where only pedestrians' trajectories are considered. The training and validation sets are combined and passed to the fitting function.

The input features of our model include the last 7 positions relative to the 8th position, with a timestep of 400 ms between each two positions. These points are rotated by the latest heading angle, and we also consider the sum of lost, occluded, and generated flags for the eight input points. Finally, the width and height of the object at the 8th point are included as well.

The target variables consist of the $(x,y)$ coordinates of the object after 4.8 seconds, rotated and transferred in a manner similar to the input trajectory case. This requires two \ac{EBM} models for each coordinate.

\begin{figure}[h]
    \centering
    \includegraphics[width=0.8\textwidth]{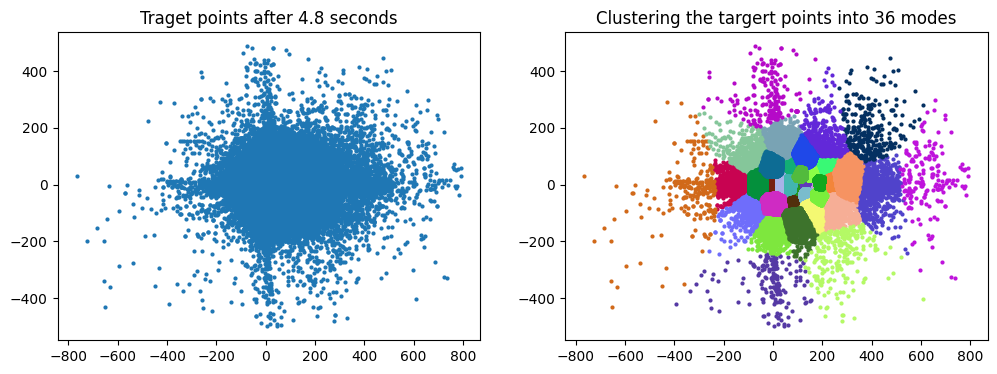}
    \caption{Left: full training set of target variable. Right: target variable set is split into multiple clusters.}
    \label{fig:sdd_traget}
\end{figure}

\vspace*{-\baselineskip}

\Cref{fig:sdd_traget} on the left shows the target variable for both training and validation sets of pedestrians combined. The data is concentrated towards the centre, but there are some points that are far away due to noise in SDD as mentioned earlier (\cite{trajnet}). To address this issue, we removed any point outside a specific range for both $x$ and $y$, which represents only 0.005\% of the total data.

In previous works like \cite{ynet,TDOR,p2tirl}, evaluations typically involved using 20 modes as outputs. However, since we started by clustering the data into 36 groups using k-means algorithm (as shown in \cref{fig:sdd_traget} on the right), and then trained two separate models for each $x$ and $y$ within these clusters, only the most probable 20 pairs of predictions  will be selected. Furthermore, we also trained global uni-modal models to help identify the best models later in the process.




\vspace*{-\baselineskip}
\subsection{InD}


For each scene, the data from its CSV file is read, and then we formulate the input similarly to \ac{SDD}, as in previous studies such as \cite{acvrnn,goalsar}. This includes the last 8 positions of the past 3.2 seconds, which consist of their position, velocity, acceleration, heading, width, and height. The target variable is the traffic entity's position after 4.8 seconds; however, only pedestrian predictions are considered in this case. Similar to \ac{SDD}, we split the target variable data into 50 modes. 


\subsection{Argoverse}

\subsubsection{Preprocessing Steps}

For Argoverse \cite{argo}, we aim to represent all available input data, including the road network and other nearby traffic entities. However, before feeding this information into our EBM model, some preprocessing is required to make it easier for the model while also maintaining interpretability of the features.

We then split the dataset based on the type of ego traffic entity present in each scene. This results in different sets of models being trained for various types of entities (cars, bicycles, buses, pedestrians, and motorcyclists), with approximately 88\% of the data representing cars while the rest is distributed among these other categories.

\vspace*{-\baselineskip}
\subsubsection{Road Network Data}

\begin{figure}[h]
    \centering
    \includegraphics[width=0.6\textwidth]{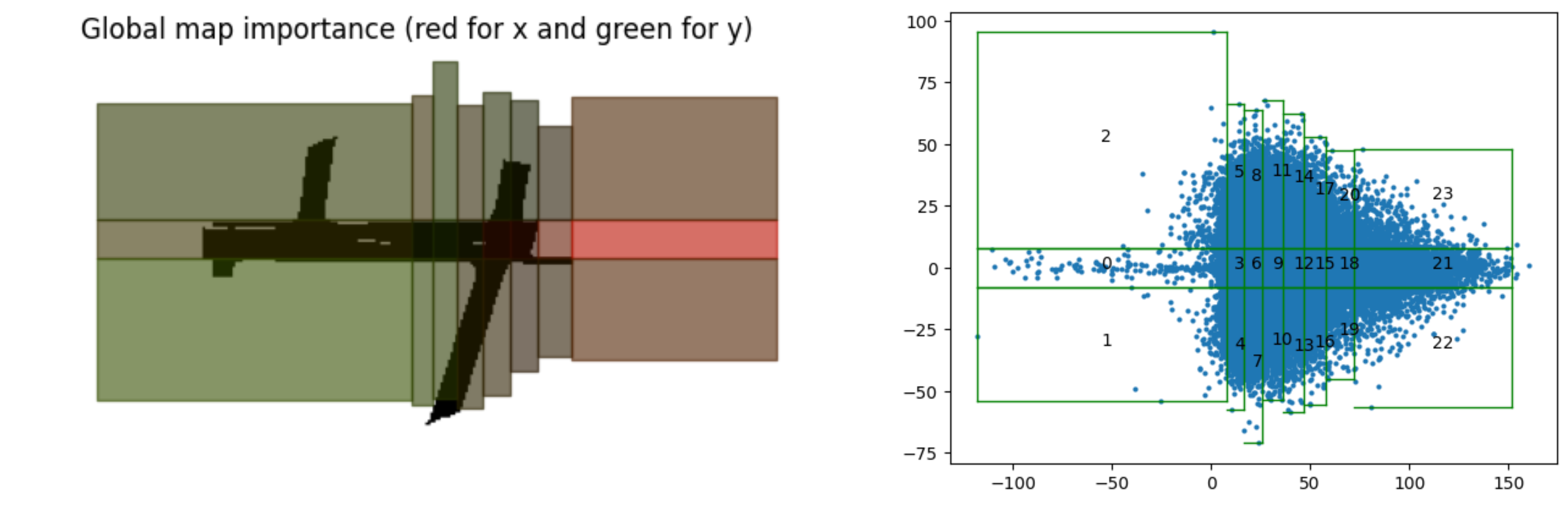}
    \caption{How the road network is represented (Left). Each mode rectangle (Right) return the geometric center of the drivable area underneath it. Red and green colors represent importance for x and y respectively}
    \label{fig:road_and_mode}
\end{figure}

The road network is initially represented as a binary image, where the drivable area has high value. Next, different sub-areas representing various modes are created based on the target variable (shown in \cref{fig:road_and_mode}). These mask parts of the drivable area and find the geometric centre of each mode rectangle. However, a tuple of zeros will be returned if there is no drivable area available.

Using these same sub-area modes from \cref{fig:road_and_mode} is an appropriate choice because they represent the most relevant part of the road for its corresponding prediction. Finally, the point of collision (POC) based on Post encroachment time (PET) (\cite{ttcpet}) is included as input features to capture traffic movement in the area. A detailed explanation of this calculation can be found in \cref{app:a}.

\subsubsection{Defining Modes}

Similar to SDD and InD, we rotate all points and move them to the last input directions and positions for of their respective trajectory. For the challenge conditions, 5 seconds of trajectory history are required as input, while a prediction is made for 6 seconds into the future. To achieve this, we define 24 modes on the target variable (shown in \cref{fig:road_and_mode} for vehicles) by splitting uniformly along the x-axis and manually dividing along the y-axis to create large central modes. Probabilities are calculated similarly to SDD and InD, with the top 6 mode predictions being selected as final.

\vspace*{-\baselineskip}
\section{Experimental Evaluation}

After training all models, we evaluate their performance using final displacement errors displayed in \cref{tab3}. The additive models show competitive results on SDD and are comparable to state-of-the-art (SoTA) on InD without road map input. However, they do not perform as well on Argoverse. One factor for that is that cars predication is a harder problem than pedestrians' predication and rely heavily on nearby elements and roadmap.

\vspace*{-\baselineskip}

\begin{table}[!htbp]
\caption{Minimum final displacement errors for 20 modes on SDD and InD, and for 6 modes on Argoverse for EBM (ours) and a subset of SoTA previous methods}
\label{tab3}
\begin{center}
\tabcolsep=0.02cm
\begin{footnotesize}
\begin{tabular}{|l|c|c|c|c|c|c|c|c|c|c|}

\hline
\bf Dataset  & \bf EBM &SEPT &ST-GAT &AC-VRNN &S-GAN &Y-Net &GOAL-SAR &P2T &PEC &TDOR
\\ \hline 
\ac{SDD} (pixels) & 14.70 & - & - & - & 41.44 & 11.85 & 11.83 & 14.08 & 15.88 & \bf 10.46  \\ \hline 
\ac{InD} & \bf 0.54 & -          & 1 & 0.80 & 0.99 & 0.56 & \bf 0.54 & - & - & -  \\ \hline 
Argoverse* & 3.88 & \bf 1.09      & - & - & - & - & - & - & - & - \\ 
\hline 
\multicolumn{11}{l}{\footnotesize{*Cars only predication. \href{https://eval.ai/web/challenges/challenge-page/1719/leaderboard/4098}{Updated leaderboard here}}} \\
\multicolumn{11}{l}{\footnotesize{SEPT: \cite{argo-sept}, ST-GAT: \cite{stgat},AC-VRNN: \cite{acvrnn}}}\\

\multicolumn{11}{l}{\footnotesize{S-GAN: \cite{socialgan},Y-Net: \cite{ynet},GOAL-SAR: \cite{goalsar}}}\\

\multicolumn{11}{l}{\footnotesize{P2T:  \cite{p2tirl},PEC: \cite{pecnet},TDOR: \cite{TDOR}}}\\

\end{tabular}
\end{footnotesize}
\end{center}
\end{table}

\vspace*{-\baselineskip}

\begin{figure}[h]
    \centering
    \includegraphics[width=0.9\textwidth]{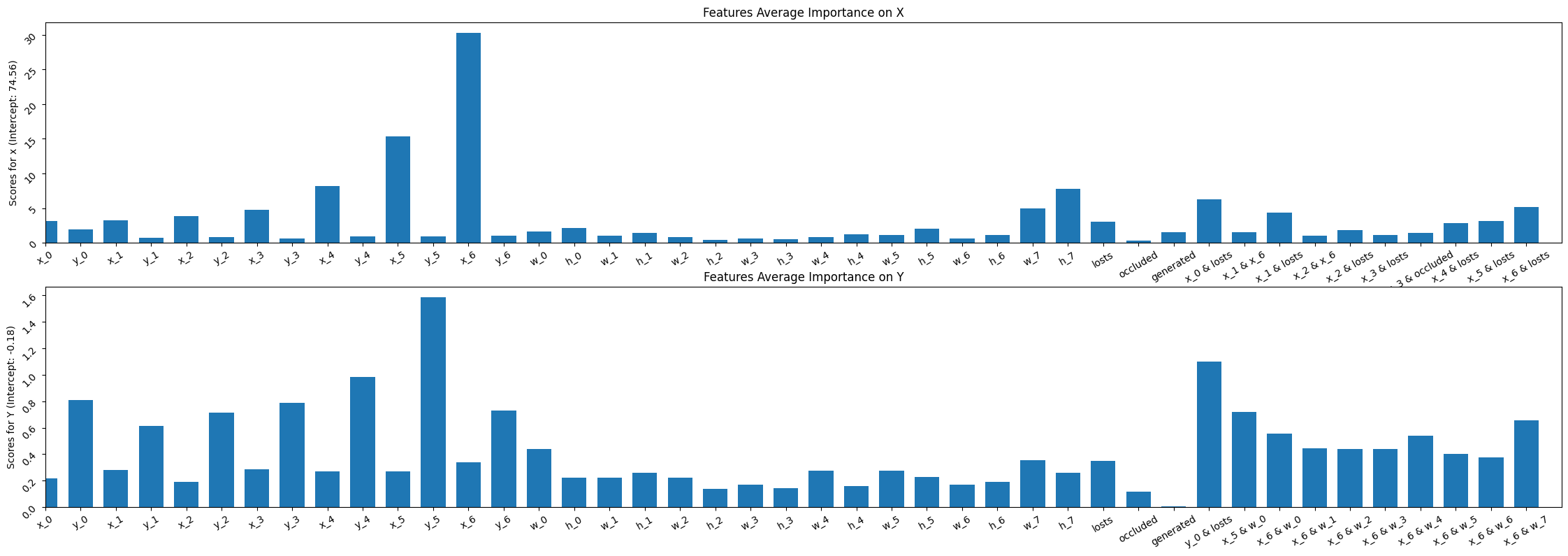}
    \caption{SDD: Global Feature Average Importance for X and Y}

    \label{fig:sdd_global}
\end{figure}

\begin{figure}[h]

    \begin{subfigure}[h]{0.5\textwidth}
    \includegraphics[width=\linewidth]{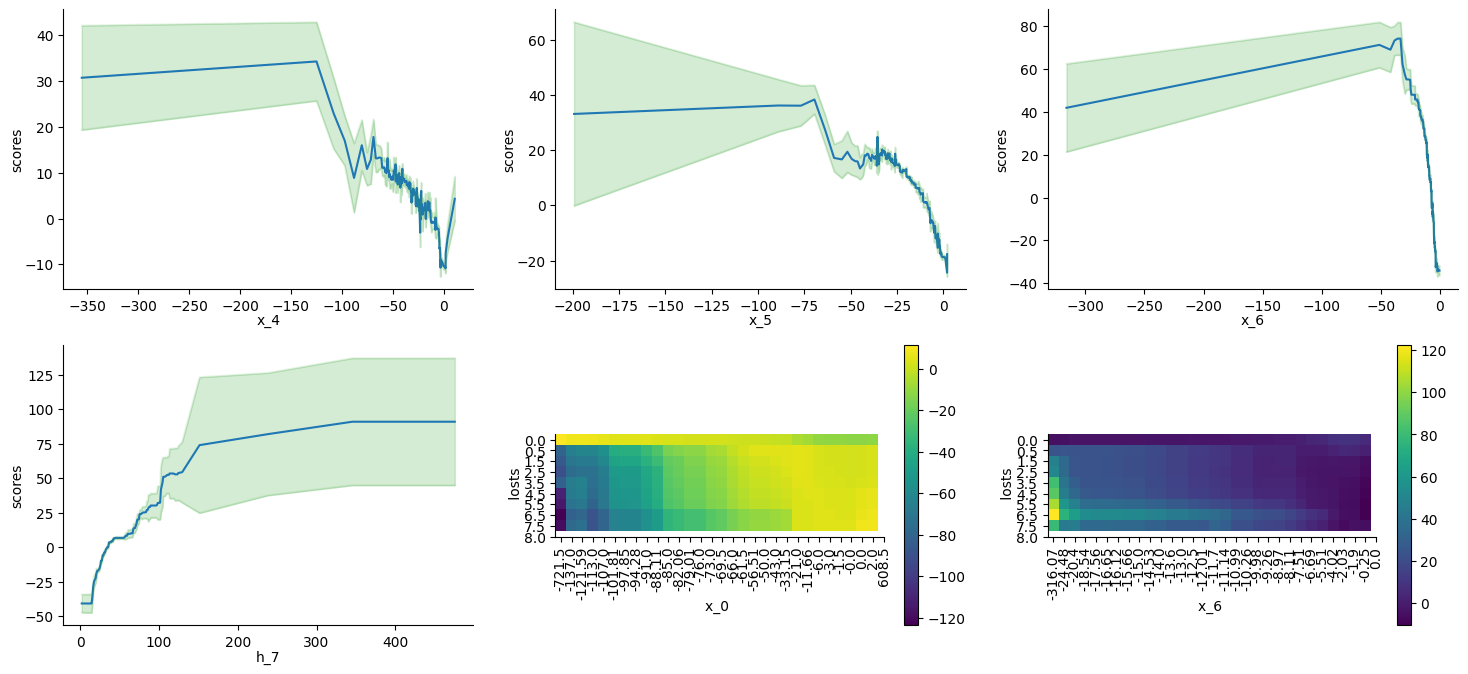}
    \caption{X-axis}
    \end{subfigure}
    \hfill
    \begin{subfigure}[h]{0.5\textwidth}
    \includegraphics[width=\linewidth]{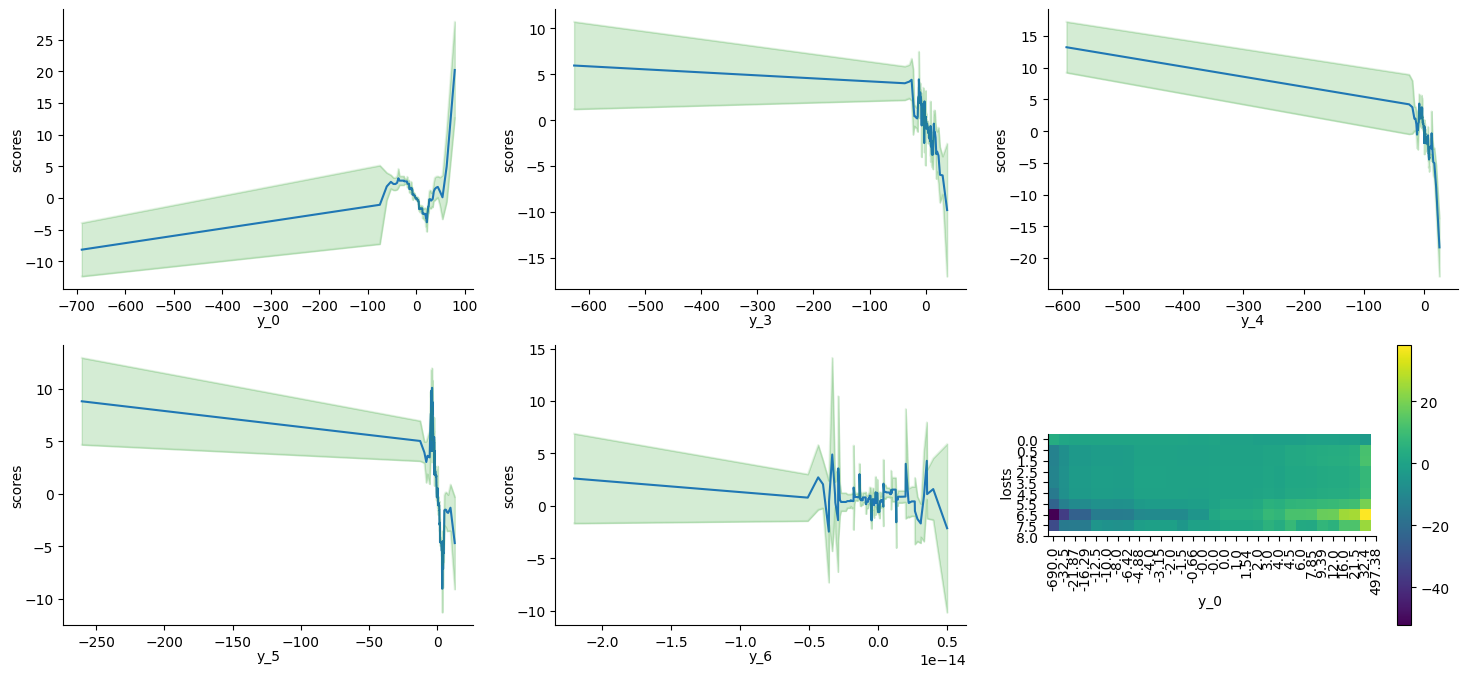}
    \caption{Y-axis}
    \end{subfigure}%
    \caption{SDD: Partial Dependence Graph on X and Y axis for the best 6 features}

    \label{fig:sdd_dg_xy}
\end{figure}

\begin{figure}[h]
    \centering
    \includegraphics[width=0.9\textwidth]{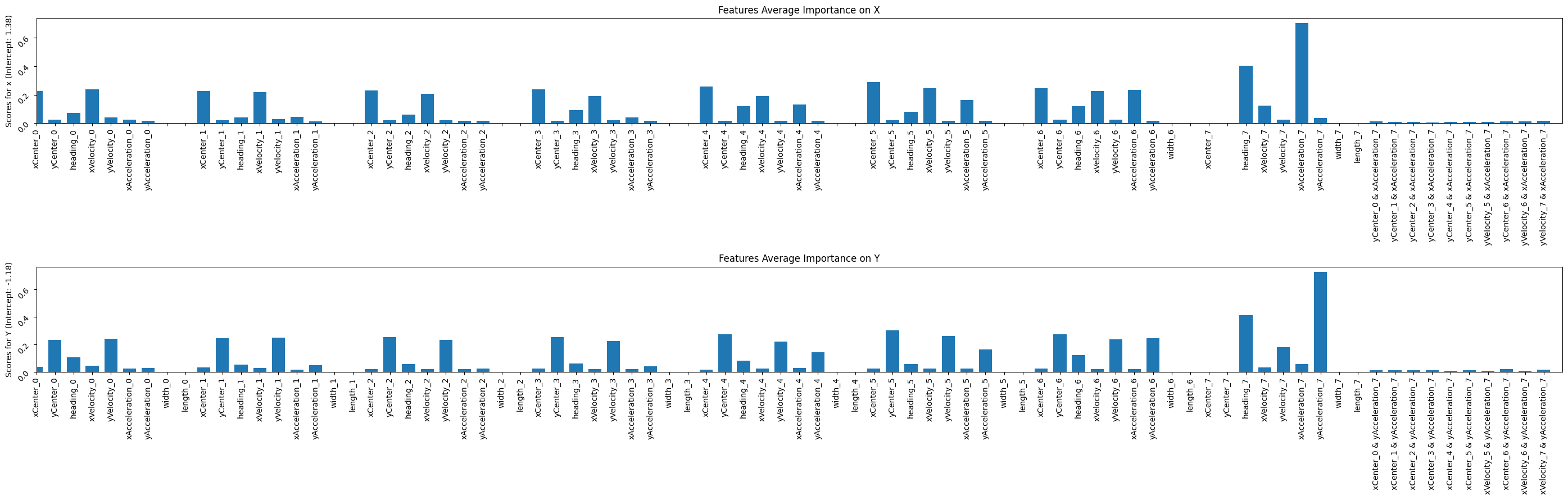}
    \caption{InD: Global Feature Average Importance for X and Y}
    \label{fig:ind_global}
\end{figure}

\begin{figure}[h]

    \begin{subfigure}[h]{0.5\textwidth}
    \includegraphics[width=\linewidth]{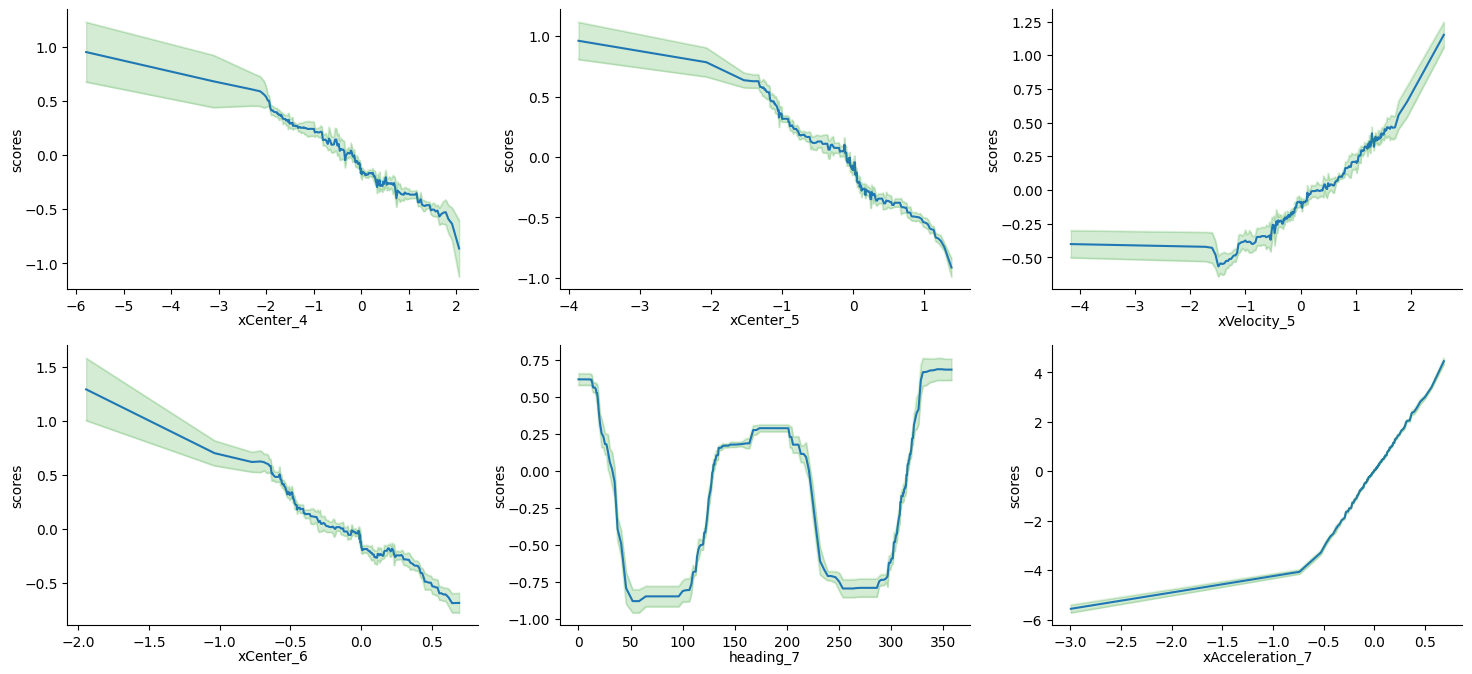}
    \caption{X-axis}
    \end{subfigure}
    \hfill
    \begin{subfigure}[h]{0.5\textwidth}
    \includegraphics[width=\linewidth]{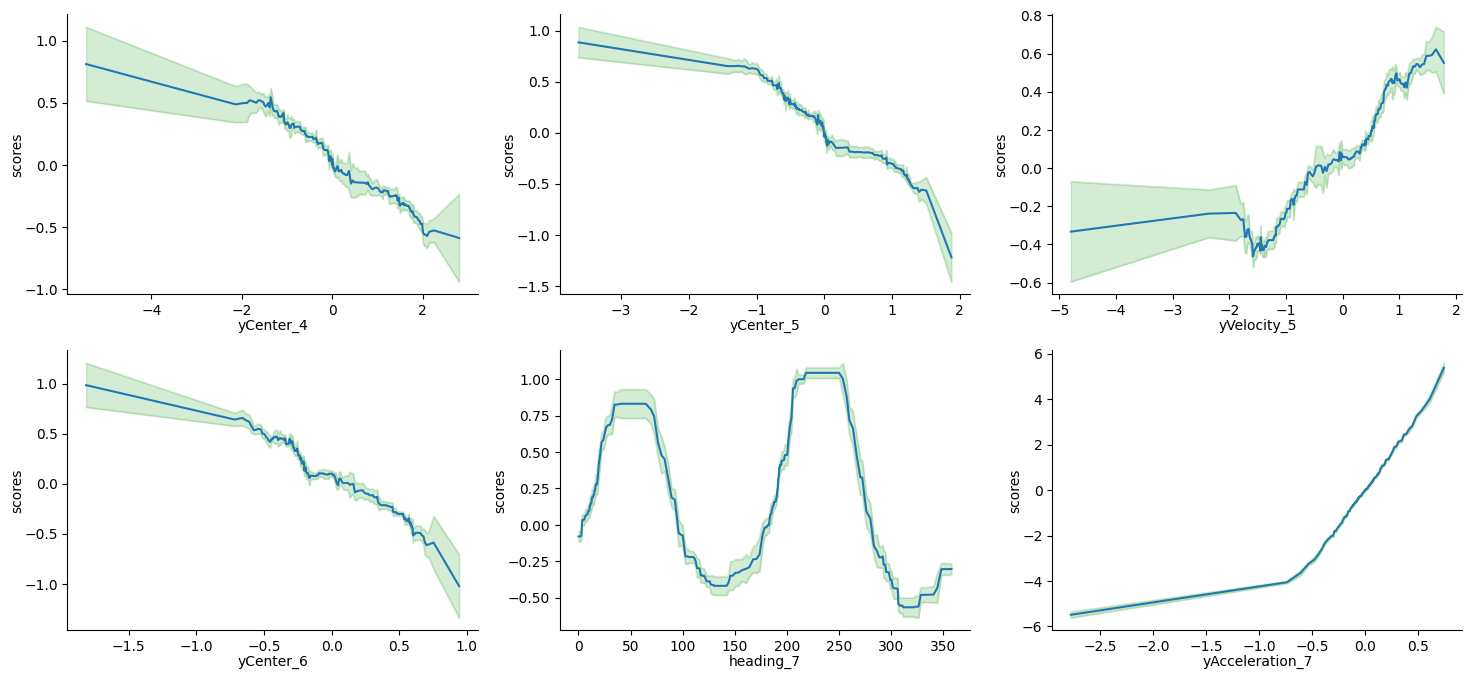}
    \caption{Y-axis}
    \end{subfigure}%
    \caption{InD: Partial Dependence Graph on X and Y axes for the best 6 features}

    \label{fig:ind_dg_xy}
\end{figure}

\begin{figure}[h]
    \centering
    \includegraphics[width=0.9\textwidth]{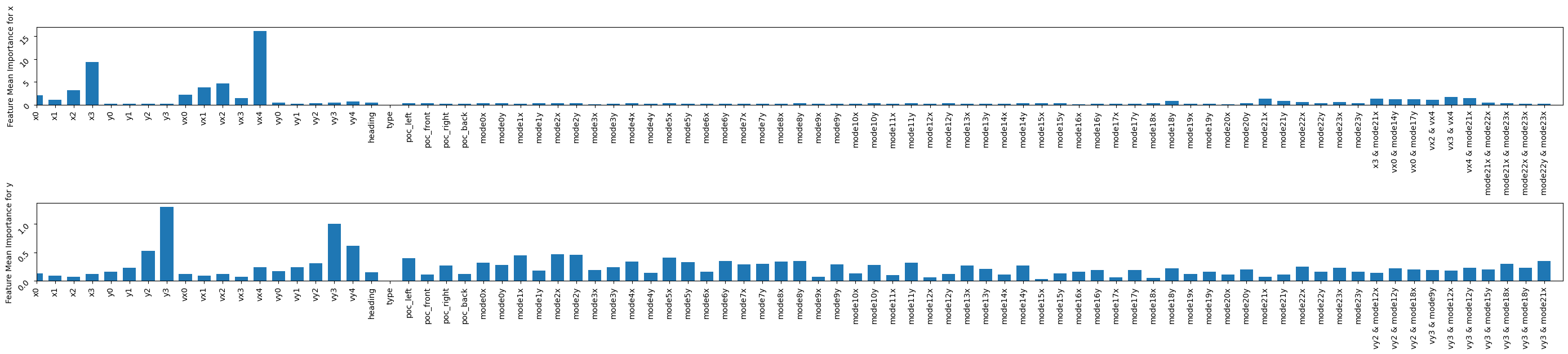}
    \caption{Argo: Global Feature Average Importance for X }
    \label{fig:argo_global}
\end{figure}

\begin{figure}[h]

    \begin{subfigure}[h]{0.5\textwidth}
    \includegraphics[width=\linewidth]{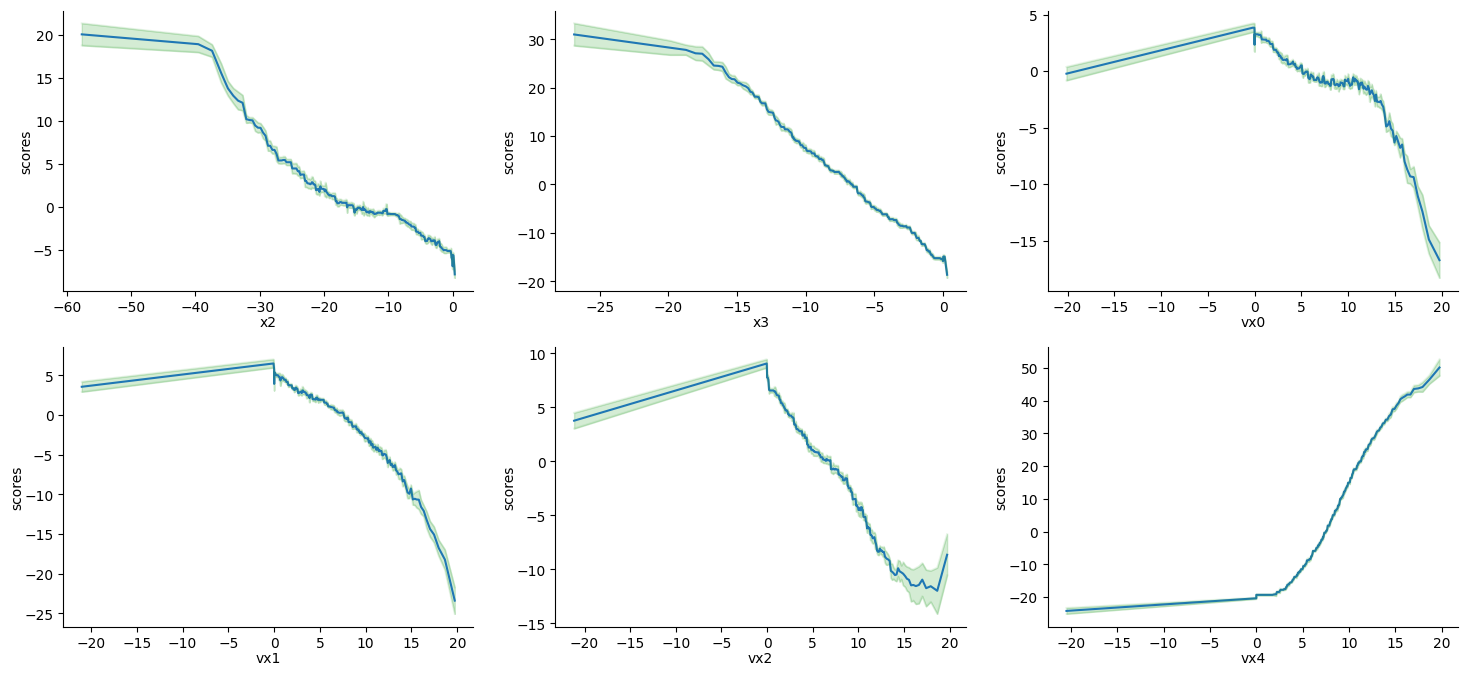}
    \caption{X-axis}
    \end{subfigure}
    \hfill
    \begin{subfigure}[h]{0.5\textwidth}
    \includegraphics[width=\linewidth]{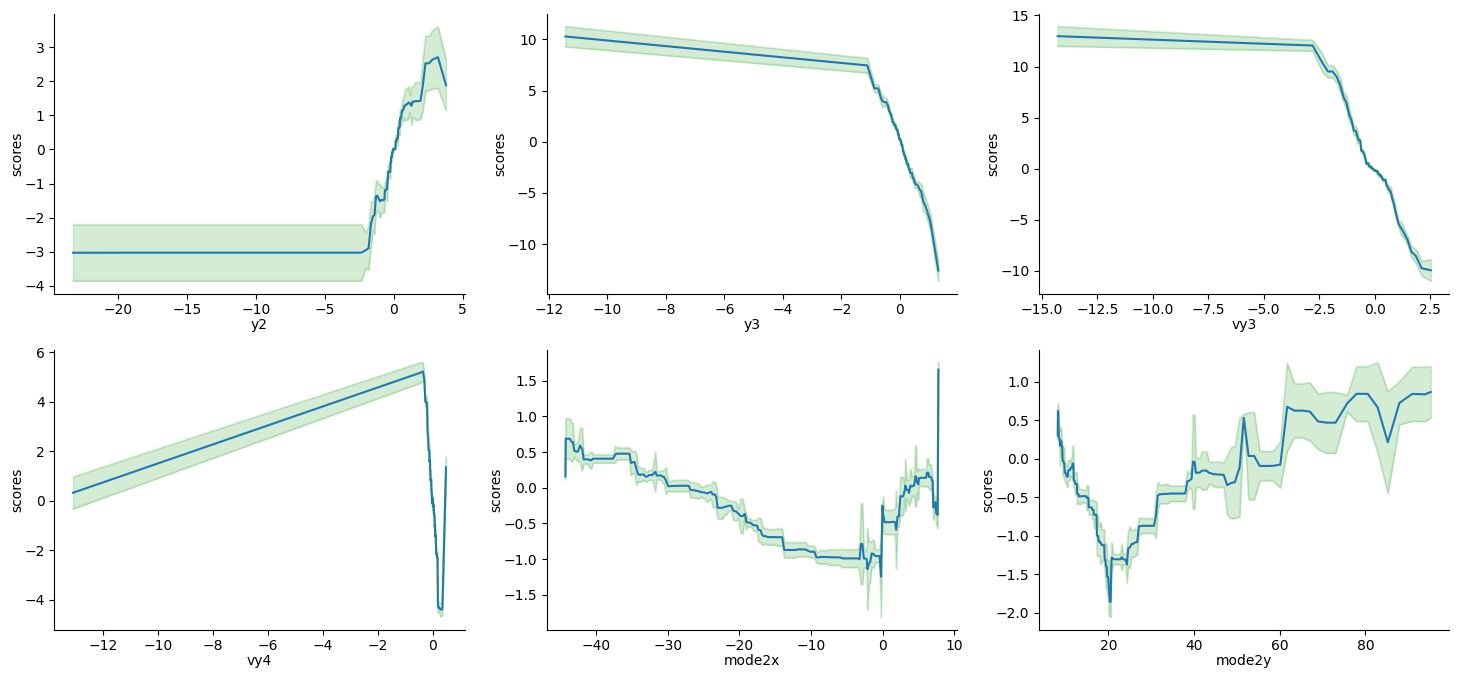}
    \caption{Y-axis}
    \end{subfigure}%
    \caption{Argo: Partial Dependence Graph on X and Y axis for the best 6 features}

    \label{fig:argo_dg_xy}
\end{figure}

\Cref{fig:sdd_global,fig:sdd_dg_xy,fig:ind_global,fig:ind_dg_xy,fig:argo_global,fig:argo_dg_xy} show global feature importance bars for each dataset using unimodal models. Additionally, partial dependence graphs of the six most important features on ($x$ and $y$) are shown. It is noted in general that the last steps are the most influencing input in all of the datasets. Another interesting remark is that the acceleration is very crucial in InD dataset models. Examples of local explanations can be found in \cref{app:c}.

\vspace*{-\baselineskip}

\section{Conclusion}

In this comprehensive analysis of generalized additive models for traffic prediction, we have seen their notable performance when it comes to predicting pedestrian movements. In fact, they outperform state-of-the-art deep learning models in terms of accuracy. Furthermore, as a glass-box model, they offer full transparency and interpretability. However, one potential drawback is their limited ability to predict multiple variables per model, which requires the training of numerous individual models for each point along the trajectory. Future research will focus on developing more effective image processing techniques for the road network, allowing us to fully leverage the environmental information available in the data.

\clearpage

\section*{Reproducibility Statement and Supplementary Material}

The full training code is split into three notebooks for the three datasets, with the instruction of how to run them here:

\url{https://anonymous.4open.science/r/GAM4Traffic-5785/Readme.md}

The full training (and testing) can be reproduced fairly fast on modest hardware. In our case, the training was done on (Intel(R) Core(TM) i7-10850H CPU)

\vskip 0.2in
\bibliography{sample}

\appendix

\section{Representing Nearby Traffic}
\label{app:a}

\begin{figure}[h]
    \centering
    \includegraphics[width=0.6\textwidth]{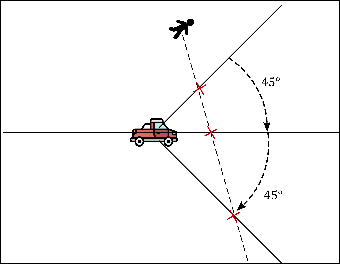}
    \caption{Finding the points of collision for four directions }
    \label{fig:poc}
\end{figure}

The future trajectory is also influenced by nearby entities, which need to be compressed for better training and model interpretability. One approach to achieve this is through direct measures related to movement such as Time to collision (TTC) or Post encroachment alignment (PET). However, both of these methods rely on time values while we require position predictions. Therefore, in this section, the point of collision (POC) will be used as a measure. POC assumes that two objects have constant velocity movement and predicts their expected point of collision. To account for all possible scenarios, four different directions of movement are assumed for the ego object, as shown in \cref{fig:poc}. The mean of other objects' POC with those directions is calculated to give one point per direction. If no point is found for a particular direction, a default mid-point will be returned.

Lastly, it should be noted that since one coordinate is redundant given the movement direction; only four values representing the x coordinates of these four points are taken as input features.

\section{EBM Hyperparameters}
\label{app:b}

List of the hyperparameters of \ac{EBM} in this work are shown in \cref{tab:ebm_hyperparams}, such as maximum number of leafs, and maximum training round, learning rate, maximum feature bins, and outer bags.

\begin{table}[h]
\centering
\caption{\ac{EBM} Hyperparameters}
\label{tab:ebm_hyperparams}
\begin{tabular}{|l|c|}
\hline
Hyperparameter & Value \\ \hline 
Maximum feature bins & 256  \\
Maximum interaction bins & 32  \\
Maximum training rounds & 5000   \\
Learning rate & 0.01   \\
Maximum number of leaf & 3 \\
Outer bags & 8  \\
Validation size & 15\%  \\
\hline
\end{tabular}
\end{table}

\section{Qualitative Examples of Local Explanation}
\label{app:c}

\begin{figure}[h]
    \centering
    \includegraphics[width=1\textwidth]{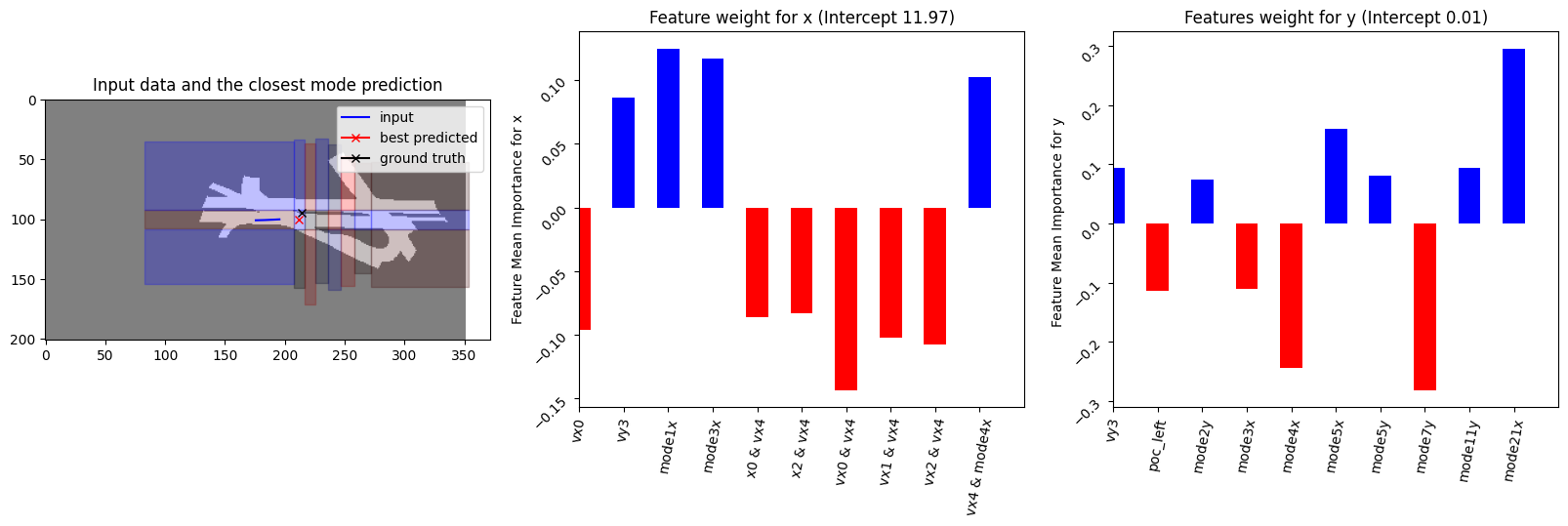}
    \caption{Argo: Local Example of the best mode in one case. Road highlighted according to X values (red increase it and blue decrease it)}
    \label{fig:argo_local_1}
\end{figure}

\begin{figure}[h]
    \centering
    \includegraphics[width=1\textwidth]{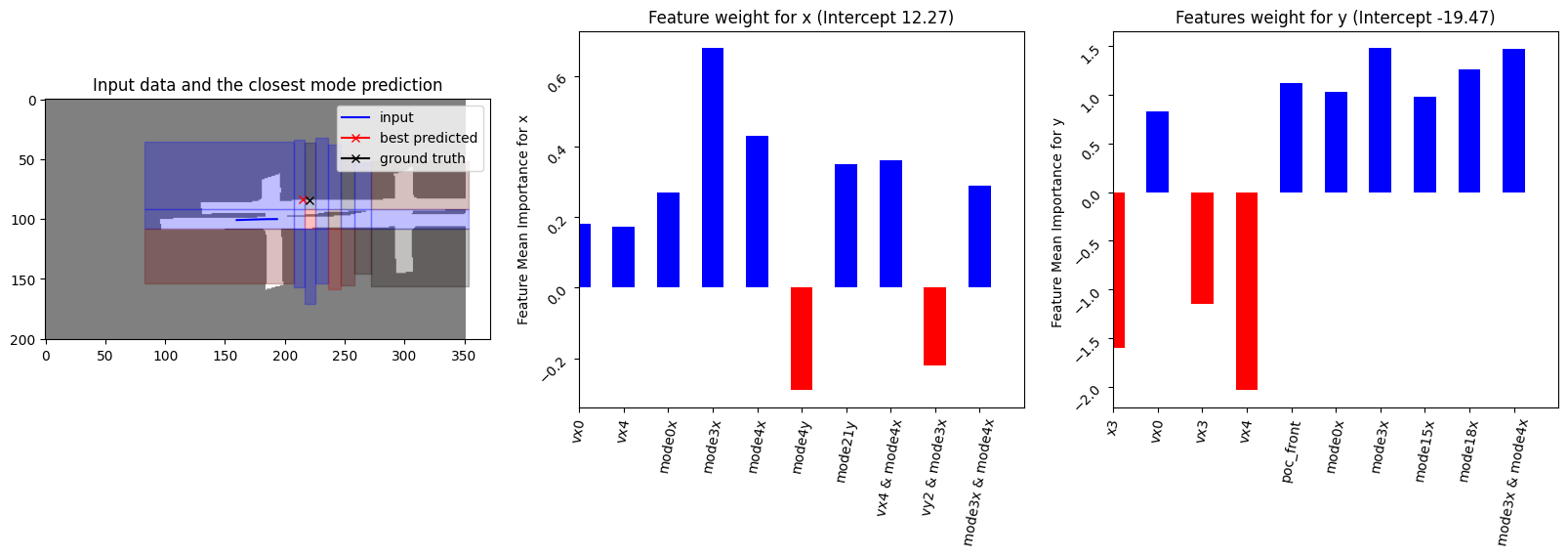}
    \caption{Argo: Local Example of the best mode in one case. Road highlighted according to X values (red increase it and blue decrease it)}
    \label{fig:argo_local_2}
\end{figure}

\begin{figure}[h]
    \centering
    \includegraphics[width=1\textwidth]{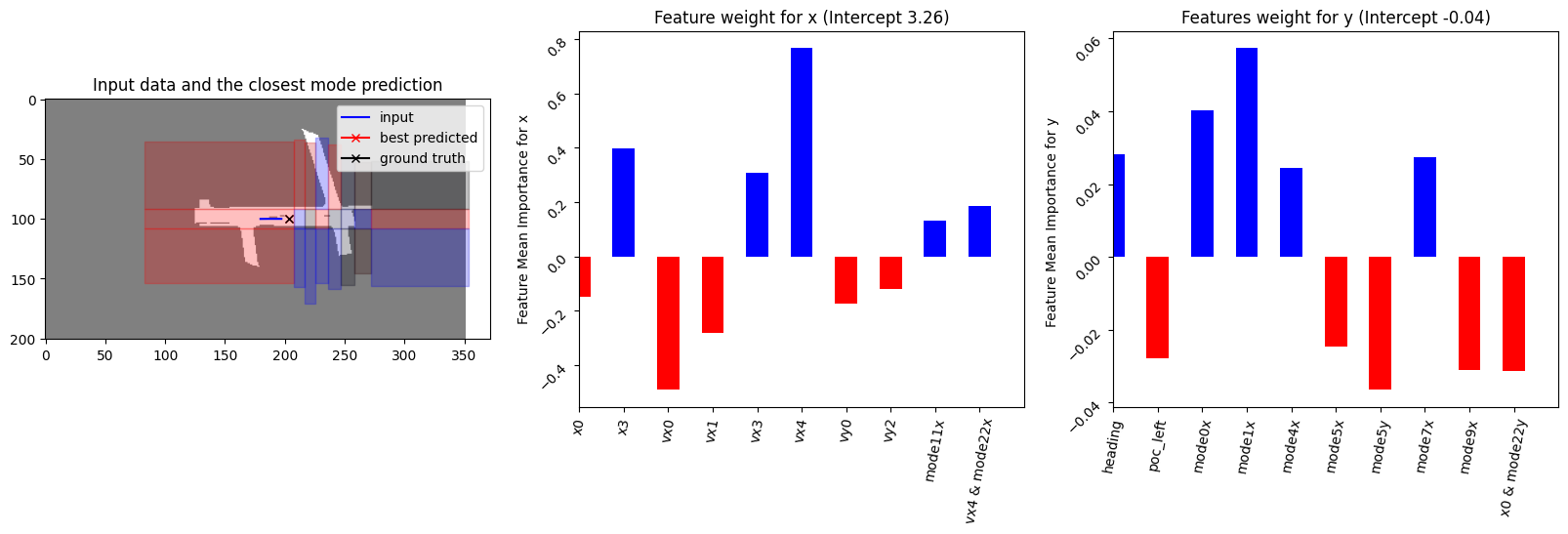}
    \caption{Argo: Local Example of the best mode in one case. Road highlighted according to X values (red increase it and blue decrease it)}
    \label{fig:argo_local_3}
\end{figure}

\begin{figure}[h]
    \centering
    \includegraphics[width=1\textwidth]{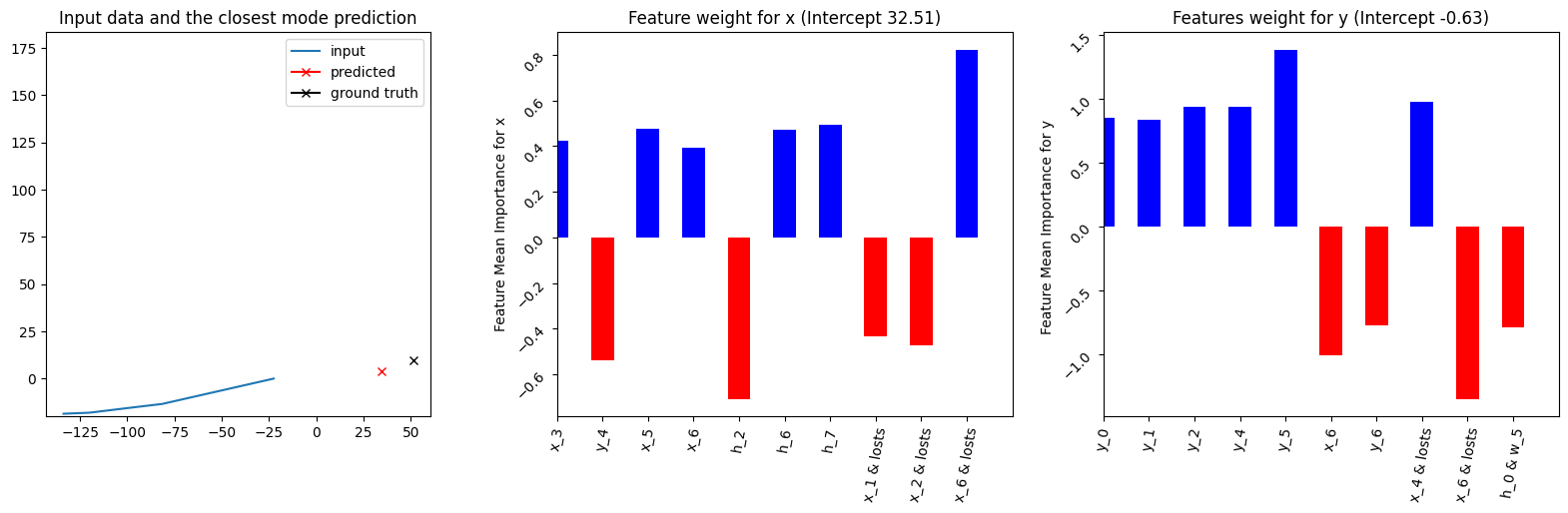}
    \caption{SDD: Local Example for one point prediction}
    \label{fig:sdd_local_1}
\end{figure}

\begin{figure}[h]
    \centering
    \includegraphics[width=1\textwidth]{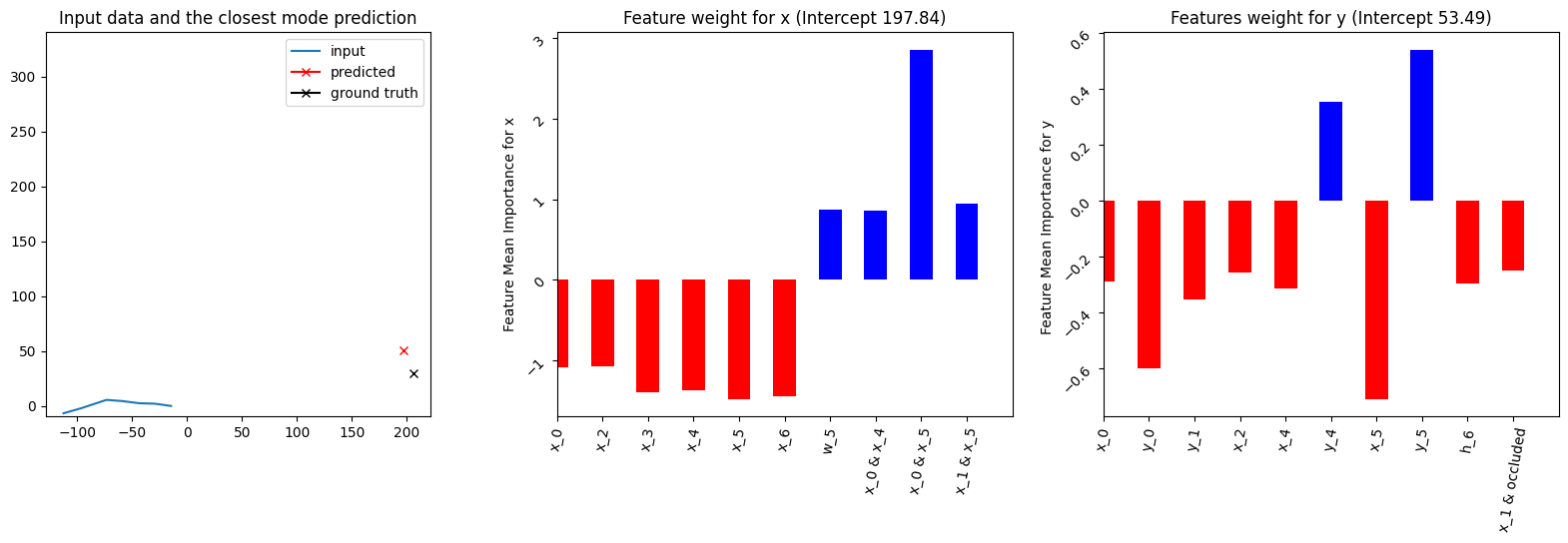}
    \caption{SDD: Local Example for one point prediction}
    \label{fig:sdd_local_2}
\end{figure}

\begin{figure}[h]
    \centering
    \includegraphics[width=1\textwidth]{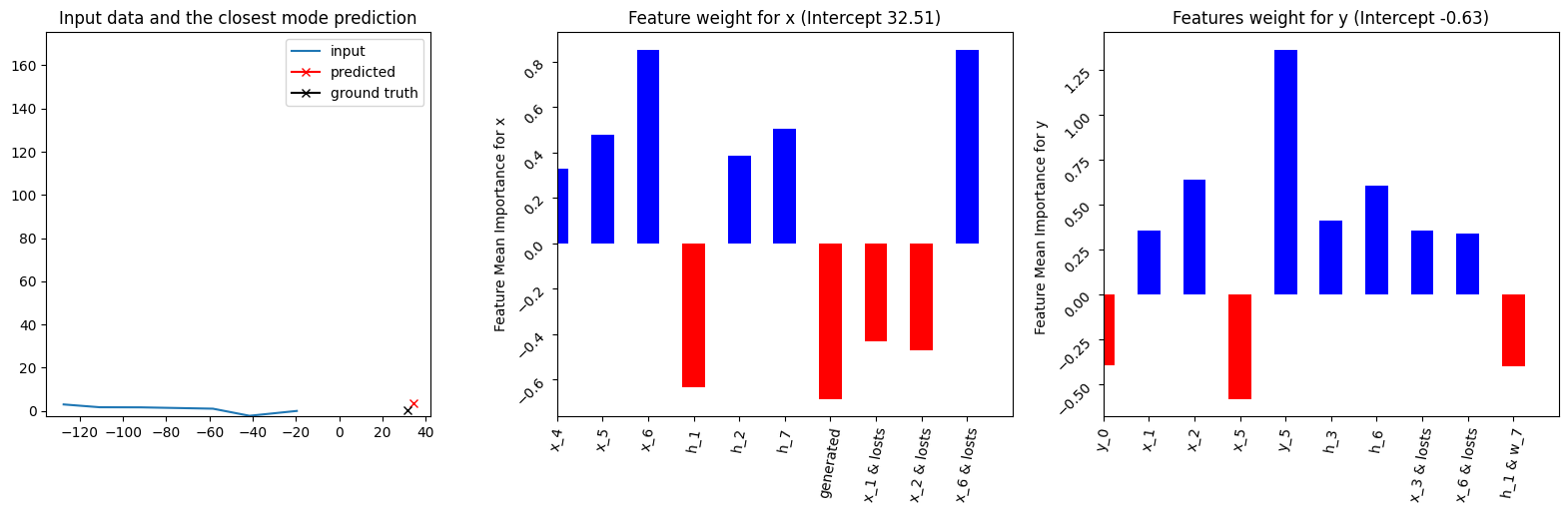}
    \caption{SDD: Local Example for one point prediction}
    \label{fig:sdd_local_3}
\end{figure}

\begin{figure}[h]
    \centering
    \includegraphics[width=1\textwidth]{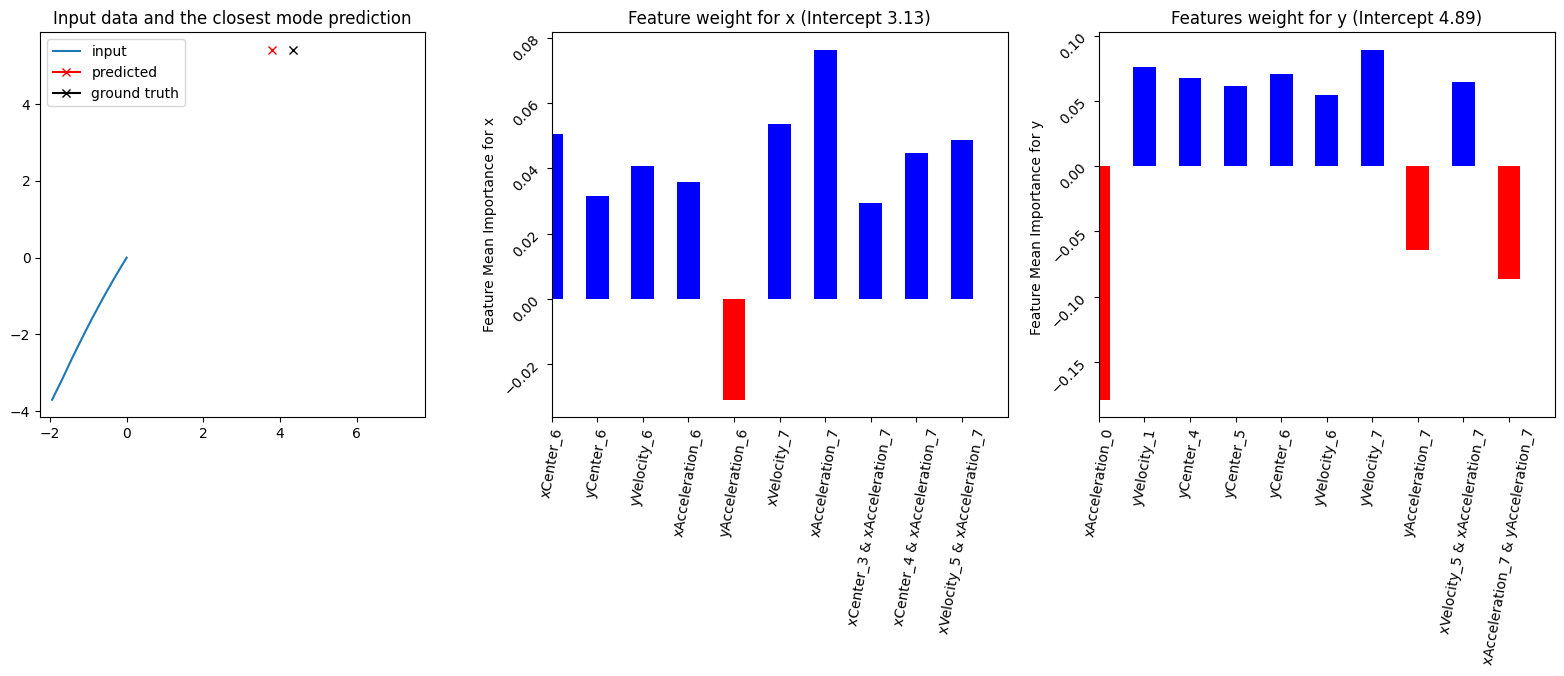}
    \caption{InD: Local Example for one point prediction}
    \label{fig:ind_local_1}
\end{figure}

\end{document}